\documentclass{article}
\usepackage{ijcai22}

\usepackage{times}
\usepackage{soul}
\usepackage{url}
\usepackage[hidelinks]{hyperref}
\usepackage[utf8]{inputenc}
\usepackage[small]{caption}
\usepackage{graphicx}
\usepackage{amsmath}
\usepackage{amssymb}
\usepackage{amsthm}
\usepackage{booktabs}
\usepackage{algorithm}
\usepackage{algorithmic}
\usepackage{bbm}
\usepackage{multirow}
\usepackage{color}
\usepackage{subfigure}
\newcommand{\hc}[1]{{\color{black} #1}}
\usepackage{pdfpages}
\usepackage{dsfont}

\title{SimMC: Simple Masked Contrastive Learning of Skeleton Representations for Unsupervised Person Re-Identification}

\author{
Haocong Rao\And
Chunyan Miao\footnotemark[2]
\affiliations
School of Computer Science and Engineering, Nanyang Technological University\\
Joint NTU-UBC Research Centre of Excellence in Active Living for the Elderly (LILY)\\
\emails
haocong001@ntu.edu.sg,
ascymiao@ntu.edu.sg
}

\begin{document}

\maketitle
\footnotetext[2]{Corresponding author}

\begin{abstract}
Recent advances in skeleton-based person re-identification (re-ID) obtain impressive performance via either hand-crafted skeleton descriptors or skeleton representation learning with deep learning paradigms. However, they typically require skeletal pre-modeling and label information for training, which leads to limited applicability of these methods. In this paper, we focus on \textit{unsupervised} skeleton-based person re-ID, and present a generic Simple Masked Contrastive learning (SimMC) framework to learn effective representations from \textit{unlabeled} 3D skeletons for person re-ID. Specifically, to fully \hc{exploit} skeleton features within each skeleton sequence, we first devise a \textit{masked prototype contrastive learning (MPC)} scheme to cluster the most typical skeleton features (\hc{\textit{skeleton prototypes}}) from different subsequences randomly masked from raw sequences, and contrast the inherent similarity between skeleton features and different prototypes to learn discriminative skeleton representations without using any label. Then, considering that different subsequences within the same sequence usually enjoy strong correlations due to the nature of motion continuity, we propose the \textit{masked intra-sequence contrastive learning (MIC)} to capture intra-sequence pattern consistency between subsequences, so as to encourage learning more effective skeleton representations for person re-ID. Extensive experiments validate that the proposed SimMC outperforms most state-of-the-art skeleton-based methods. We further show its scalability and efficiency in enhancing the performance of existing models. Our codes are available at \href{https://github.com/Kali-Hac/SimMC}{https://github.com/Kali-Hac/SimMC.}

\end{abstract}

\section{Introduction}
Person re-identification (re-ID) targets at retrieving and matching the same pedestrian from different views or occasions, which assumes a pivotal role in various applications such as intelligent surveillance, robotics, and security authentication
\cite{ye2021deep}. Recently, person re-ID via 3D skeletons has drawn \hc{growing interests} from academia and industry \cite{pala2019enhanced,rao2021self,rao2021multi}. Compared with conventional image-based methods that typically rely on visual features such as human silhouettes and appearances for recognition \cite{liu2015enhancing}, skeleton-based methods leverage 3D positions of key body joints to characterize discriminative structural and motion features of human body, which could enjoy smaller data size and better robustness against scale and view variation \cite{han2017space}. 

\begin{figure}
    \centering
    \scalebox{0.37}{
    \includegraphics{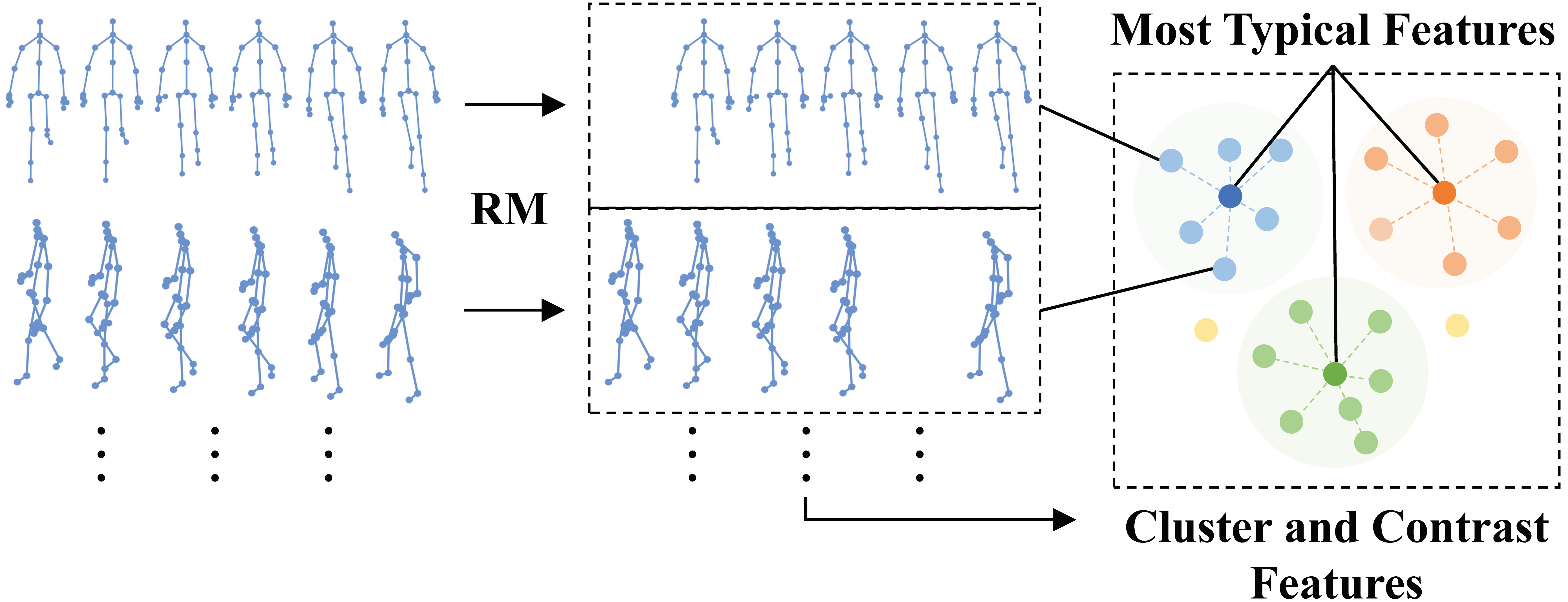}
    }
    \caption{Our framework clusters the randomly masked (RM) skeleton sequences, and contrasts their features with the most typical ones to learn discriminative skeleton representations for person re-ID.}
    \label{first}
\end{figure}

Despite the great progress in skeleton-based person re-ID, existing endeavors require either extracting hand-crafted features ($e.g.,$ anthropometric attributes) \cite{pala2019enhanced} or learning skeleton representations with the supervision of labels. For hand-crafted methods, they typically require extensive domain knowledge while lacking the flexibility to explore latent features beyond human cognition. To tackle this issue, numerous recent works resort to convolutional neural networks (CNN) \cite{liao2020model} and long short-term memory (LSTM) \cite{rao2021sm} to perform supervised or self-supervised skeleton representation learning. However, these methods usually require a specific pre-modeling of 3D skeletons ($e.g.,$ skeleton graphs \cite{rao2021multi}), and rely on massive manually-annotated data to train or fine-tune models, which is labor-expensive and unable to learn general pedestrian representations under the unavailability of labels.

To address these challenges, this paper presents a generic Simple Masked Contrastive learning (SimMC) framework, as shown in Fig. \ref{first}, which contrasts the typical features and inherent relationships of \textit{masked} skeleton sequences to learn effective skeleton representations \textit{without using any label} for person re-ID. Specifically, to fully \hc{utilize} unique features within skeleton sequences, we first devise a \textbf{\textit{masked prototype contrastive learning (MPC)}} scheme to cluster \textit{subsequence} representations (referred as \textit{skeleton instances}) randomly masked from raw sequences, and contrast the inherent similarity between them and the most typical features (referred as \textit{skeleton prototypes}) to learn discriminative skeleton representations. By pulling closer skeleton instances belonging to the same prototype and pushing apart instances of different prototypes with the instance-prototype contrastive learning, MPC enables the model to capture discriminative skeleton features and high-level semantics ($e.g.,$ intra-class skeleton similarity) from \textit{unlabeled} skeleton sequences for the person re-ID task. Then, motivated by the nature of motion continuity that typically endows different subsequences with strong correlations ($e.g.,$ motion similarity), we propose the \textbf{\textit{masked intra-sequence contrastive learning (MIC)}} to learn the intra-sequence similarity between subsequences of the same skeleton sequence, which encourages capturing the pattern consistency within sequences to learn more effective representations of skeletons for person re-ID. 

The proposed SimMC framework enjoys merits in terms of architectures, performance, and scalability. Firstly, SimMC is primarily built by multi-layer perceptron (MLP) networks with small model complexity, which can directly learn effective representations from raw skeleton sequences without any prior modeling. 
Secondly, the proposed unsupervised framework outperforms most existing self-supervised and supervised skeleton-based methods that utilize extra label information, and can also be efficiently applied to 3D skeleton data estimated from RGB-based scenes. Lastly, our framework can serve as a generic contrastive learning paradigm to fine-tune skeleton features learned from existing models, which benefits learning better skeleton representations for the task of person re-ID. 
In summary, our main contributions include:

\begin{itemize}
    \item We present a simple masked contrastive learning (SimMC) framework that exploits typical features and relationships of masked unlabeled skeleton sequences to learn discriminative representations for person re-ID. 
    
    \item We devise a novel masked prototype contrastive learning (MPC) scheme to fully contrast most representative features and learn high-level semantics from subsequence representations masked from skeleton sequences.
    
    
    \item We propose the masked intra-sequence contrastive learning (MIC) to learn inherent similarity and pattern consistency between subsequences, so as to encourage learning more effective representations for person re-ID. 
    
    \item Empirical evaluations show that SimMC significantly outperforms most state-of-the-art skeleton-based methods on four benchmark datasets, and can be exploited to fine-tune existing skeleton representations and boost their performance with up to $28.2\%$ mAP gains.
\end{itemize}


\section{Related Works}
\paragraph{Skeleton-based Person Re-identification.} 
Most existing methods typically extract hand-crafted anthropometric, morphological, and gait descriptors from 3D skeletons to characterize human body and motion features. 
\hc{Seven Euclidean distances between certain joints are utilized by \cite{barbosa2012re} to construct a distance matrix for person re-ID. Further enhancement with 13 ($D_{13}$) and 16 skeleton descriptors ($D_{16}$) are made in 
 \cite{munaro2014one} and \cite{pala2019enhanced}, respectively, which leverage $k$-nearest neighbor, support vector machine or Adaboost classifiers to perform person re-ID. }
Recently, deep neural networks are widely applied to supervised and self-supervised skeleton representation learning.
\hc{A CNN-based paradigm, PoseGait \cite{liao2020model}, is devised to encode 81 hand-crafted skeleton/pose features for human recognition. An LSTM-based skeleton encoding model with locality-aware attention (AGE) \cite{rao2020self} is proposed to learn discriminative gait features from skeleton sequences.}
SGELA \cite{rao2021self} further combines multiple self-supervised pretext tasks ($e.g.,$ reverse sequential reconstruction) and inter-sequence contrastive scheme to enhance skeleton pattern learning for person re-ID. The graph-based methods MG-SCR \cite{rao2021multi} and SM-SGE \cite{rao2021sm} devise multi-level skeleton graphs and auxiliary self-supervised tasks for person re-ID tasks.

\paragraph{Contrastive Learning.}
Contrastive learning is widely applied to various self-supervised and unsupervised paradigms \cite{he2020momentum,rao2021self,rao2021augmented,chen2021exploring} to learn effective data representations by pulling together positive representation pairs and pushing apart negative ones in a certain feature space. \hc{An instance discrimination paradigm based on exemplar tasks \cite{wu2018unsupervised} is devised for image contrastive learning. The contrastive predictive coding (CPC) model with the probabilistic InfoNCE loss \cite{oord2018representation} is proposed to learn general representations from various domains.} Recent contrastive paradigms explore mini-batch negative sampling \cite{chen2020a} and momentum-based encoders \cite{he2020momentum}, while \cite{chen2021exploring} devises a Siamese architecture for contrastive learning without using negative pairs or momentum encoders.
In \cite{li2021prototypical}, contrastive learning and \textit{k}-means clustering are combined for unsupervised learning of visual representations.

\section{The Proposed Framework}
\label{approach}
\begin{figure}
    \centering
    \scalebox{0.61}{
    \includegraphics{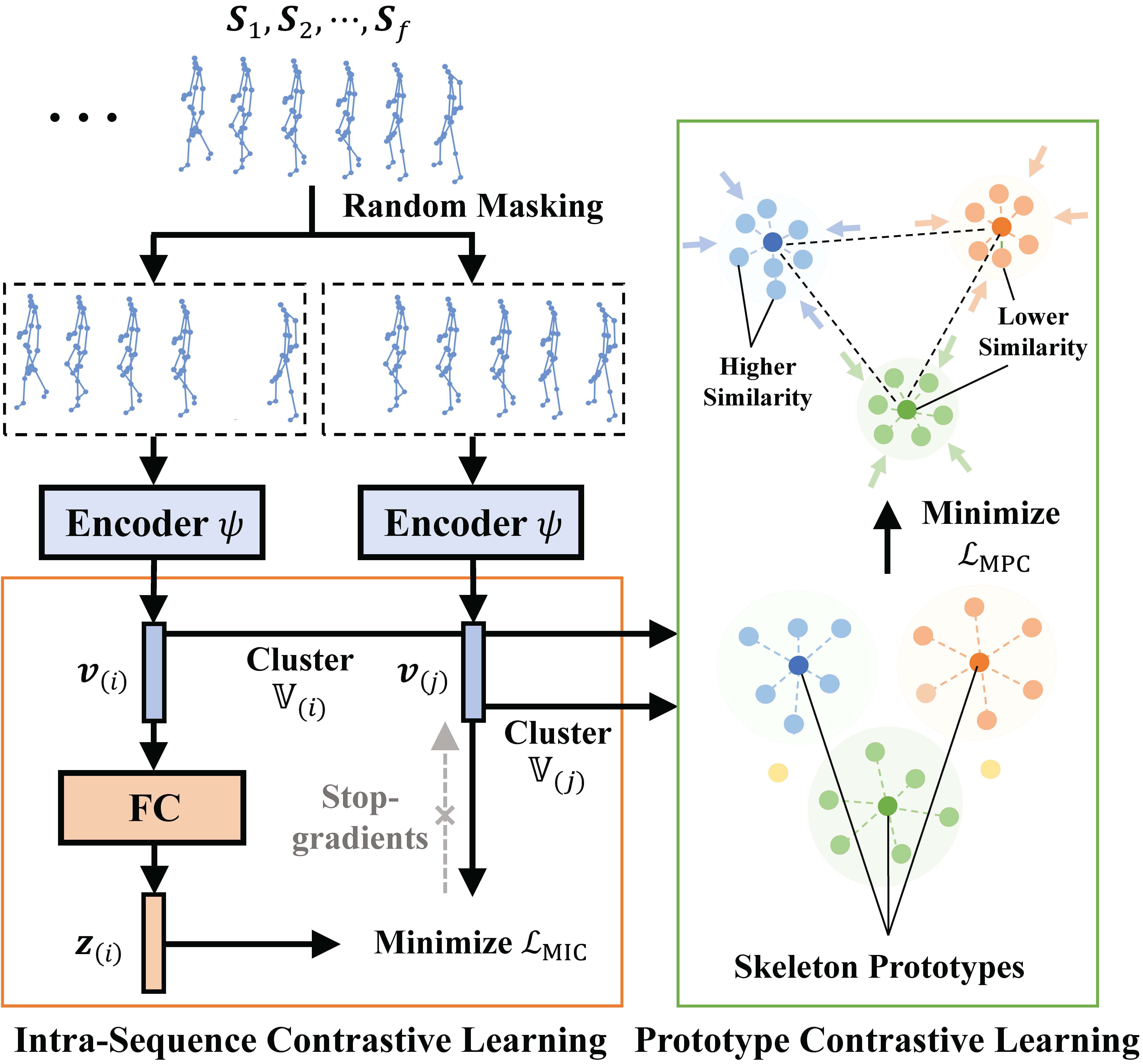}
    }
    \caption{\hc{Schematics of our framework with masked prototype contrastive learning and masked intra-sequence contrastive learning.} }
    \label{model}
\end{figure}

Suppose that a 3D skeleton sequence $\boldsymbol{S}_{1:f}\!=\!(\boldsymbol{S}_1,\cdots,\boldsymbol{S}_{f})\in \mathbb{R}^{f \times K}$, where $\boldsymbol{S}_{t}\in \mathbb{R}^{K}$ is the $t^{th}$ skeleton with 3D coordinates of $J$ body joints and $K\!=\!J\times 3$. Each skeleton sequence $\boldsymbol{S}_{1:f}$ belongs to an identity $\text{y}$, where $\text{y}\in \{1, \cdots, I\}$ and $I$ is the number of different identities. 
The training set $\Phi_{\mathcal{T}}=\left\{\boldsymbol{S}^{ \mathcal{T},i}_{1:f}\right\}_{i=1}^{N_{1}}$, probe set $\Phi_{\mathcal{P}}=\left\{\boldsymbol{S}^{ \mathcal{P},i}_{1:f}\right\}_{i=1}^{N_{2}}$, and gallery set $\Phi_{\mathcal{G}}=\left\{\boldsymbol{S}^{\mathcal{G}, i}_{1:f}\right\}_{i=1}^{N_{3}}$ contain $N_{1}$, $N_{2}$, and $N_{3}$ skeleton sequences of different persons in different views and scenes. 
Our framework aims at learning an encoder (denoted as $\psi(\cdot)$) built with neural networks to encode $\Phi_{\mathcal{P}}$ and $\Phi_{\mathcal{G}}$ into effective skeleton representations $\{\boldsymbol{v}^{\mathcal{P}}_i\}_{i=1}^{N_{2}}$ and $\{\boldsymbol{\boldsymbol{v}}^{\mathcal{G}}_j\}_{j=1}^{N_{3}}$ \textit{without using any label}, such that the representation $\boldsymbol{v}^{\mathcal{P}}_i$ in probe set can match the representation $\boldsymbol{v}^{\mathcal{G}}_j$ of the same identity in gallery set. 
\hc{The overview of our framework is presented in Fig. \ref{model}.

As shown in Fig. \ref{model}, we firstly randomly mask each input skeleton sequence to sample $i^{th}$ and $j^{th}$ subsequences, which are encoded into skeleton instances $\boldsymbol{v}_{(i)}$ and $\boldsymbol{v}_{(j)}$ (see Sec. \ref{MPC_sec}). Secondly, we cluster corresponding instance sets $\mathbb{V}_{(i)}$ and $\mathbb{V}_{(j)}$ individually to generate skeleton prototypes, and then enhance the similarity between instances of same prototype while maximizing the dissimilarity between different ones by minimizing $\mathcal{L}_{\text {MPC}}$. Meanwhile, a Siamese architecture is exploited to learn inherent intra-sequence similarity between $\boldsymbol{v}_{(i)}$ and $\boldsymbol{v}_{(j)}$ by minimizing $\mathcal{L}_{\text {MIC}}$ (see Sec. \ref{MIC_sec}). }

\subsection{Masked Prototype Contrastive Learning}
\label{MPC_sec}
Each person's skeletons typically possess unique features ($e.g.,$ anthropometric attributes), while their corresponding sequences could carry recognizable and highly consistent walking patterns \cite{murray1964walking}. Naturally, we expect the model to \hc{exploit} the most representative skeleton patterns and traits \textit{within each sequence} for person re-ID. A na\"ive solution is to cluster skeleton sequences to learn the representative features by direct inter-sequence contrastive learning, while it could overlook some valuable \textit{intra-sequence} representations ($e.g.,$ subsequences) that might contain key patterns. 
To encourage the model to fully mine intra-sequence skeleton features and high-level semantics ($e.g.,$ identity-related patterns) from skeleton sequences, we propose a \textbf{\textit{masked prototype contrastive learning (MPC) scheme}} to \textit{jointly} focus on the most typical features \hc{(\textbf{\textit{skeleton prototypes}})} of different subsequence representations \hc{(\textbf{\textit{skeleton instances}})} randomly masked from original sequences, and exploit the instance-prototype similarity and dissimilarity to learn discriminative skeleton representations.

Given an input skeleton sequence $\boldsymbol{S}_{1:f}=(\boldsymbol{S}_1,\cdots,\boldsymbol{S}_{f})$, we exploit an MLP encoder with one hidden layer to encode each skeleton as:
\begin{equation}
\boldsymbol{h}_{j}=\psi\left(\boldsymbol{S}_{j}\right)=\mathbf{W}^{2} \sigma\left(\mathbf{W}^{1} \boldsymbol{S}_{j}\right),
\label{eq_1}
\end{equation}
where $\psi(\cdot)$ represents the encoder function, $\mathbf{W}^{1}\in\mathbb{R}^{H\times K}$ and $\mathbf{W}^{2}\in\mathbb{R}^{H\times H}$ denote the learnable weight matrices to encode the $j^{th}$ skeleton $\boldsymbol{S}_{j}\in\mathbb{R}^{K}$ into a latent feature representation $\boldsymbol{h}_{j}\in\mathbb{R}^{H}$,
and $\sigma(\cdot)$ is a ReLU non-linear activation function. Then, to sample subsequence representations from the encoded sequence representation $(\boldsymbol{h}_1,\cdots,\boldsymbol{h}_{f})$ of $\boldsymbol{S}_{1:f}$, we utilize a masking function $\mathcal{M}$ to randomly produce $x$ masks, $i.e.,$ zero-masking positions, for each skeleton sequence of length $f$ with:
\begin{equation}
\mathcal{M}(f, x)=(m_{1},\cdots,m_{f}), 
\label{eq_2}
\end{equation}
where $m_{j}\in\{0, 1\}$ is the mask status for the $j^{th}$ position of a sequence and $\sum^{f}_{j=1}m_{j}\!=\!f\!-\!x$. We apply the generated random masks to $\boldsymbol{S}_{1:f}$ and its corresponding skeleton representations $(\boldsymbol{h}_1,\cdots,\boldsymbol{h}_{f})$ \hc{(see Eq. (\ref{eq_1}))}, which are then integrated into a subsequence representation as (see Fig. \ref{model}): 
\begin{equation}
\boldsymbol{v}_{(i)}=\frac{1}{f-x}\sum^{f}_{j=1}m_{(i),j}w_{j}\boldsymbol{h}_{j},
\label{eq_3}
\end{equation}
where $\boldsymbol{v}_{(i)}\in\mathbb{R}^{H}$ ($i\in\{1,\cdots,q\}$) denotes the feature representation of $i^{th}$ subsequence sampled from $\boldsymbol{S}_{1:f}$ using $x$ random masks, $q$ is the number of subsequence sampling, $m_{(i),j}$ denotes the mask status of the $j^{th}$ position at the $i^{th}$ sampling, while $w_{j}$ represents the importance of $j^{th}$ skeleton representation $\boldsymbol{h}_{j}$.  Here each skeleton is assumed to equally contribute to representing 
sequence features, $i.e.$, $w_{j}=1$. For clarity, we use $\mathbb{V}_{(i)}=\{\boldsymbol{v}_{(i),j}\}_{j=1}^{N_{1}}$ to denote all subsequence representations in the $i^{th}$ subsequence sampling of the training set $\Phi_{\mathcal{T}}$. \hc{Note that} we sample one random subsequence for each training sequence at each sampling.
$\mathbb{V}_{(i)}=\{\boldsymbol{v}_{(i),j}\}_{j=1}^{N_{1}}$ are exploited as \textit{skeleton instances} for the MPC scheme. 

To group feature-similar skeleton instances and discover semantic clusters with arbitrary shapes, we leverage the DBSCAN algorithm \cite{ester1996density} to perform clustering \textit{individually} on the $i^{th}$ instance set $\mathbb{V}_{(i)}$ corresponding to $i^{th}$ subsequence sampling, as shown in Fig. \ref{model}, and generate clusters $\overline{\mathbb{V}}_{(i)}^{c}=\{\boldsymbol{v}^{c}_{(i),j}\}_{j=1}^{N_{c}}$, $c\in\{1,\cdots,C\}$, where $C$ is the number of clusters ($i.e.,$ pseudo classes), and each cluster $\overline{\mathbb{V}}_{(i)}^{c}$ contains $N_c$ instances belonging to the $c^{th}$ pseudo class. We \textit{averagely aggregate} instance features of the same cluster to generate the corresponding skeleton prototype as:
\begin{equation}
\boldsymbol{p}^{c}_{(i)}=\frac{1}{N_c}\sum^{N_c}_{j=1}\boldsymbol{v}^{c}_{(i),j} \ ,
\label{eq_4}
\end{equation}
where $\boldsymbol{p}^{c}_{(i)}\in\mathbb{R}^{H}$ denotes the skeleton prototype of the $c^{th}$ cluster $\overline{\mathbb{V}}_{(i)}^{c}$. To jointly focus on the representative skeleton features in all instance sets and encourage capturing high-level skeleton semantics from different prototypes, we exploit a masked prototype contrastive (MPC) loss to enhance the similarity of each skeleton instance to the corresponding prototype and maximize its dissimilarity to other prototypes by:
\begin{equation}
    \mathcal{L}_{\text {MPC}}=\frac{1}{N}\sum_{i=1}^{q}\sum_{c=1}^{C_{i}}\sum_{j=1}^{N_c}-\log \frac{\exp \left(\boldsymbol{v}^{c}_{(i),j} \cdot \boldsymbol{p}^{c}_{(i)} / \tau \right)}{\sum_{k=1}^{C_i} \exp \left(\boldsymbol{v}^{c}_{(i),j} \cdot \boldsymbol{p}^{k}_{(i)} / \tau \right)},
    \label{eq_5}
\end{equation}
where $N$ represents the number of all skeleton instances, $C_i$ denotes the number of skeleton prototypes generated from the $i^{th}$ instance set $\mathbb{V}_{(i)}$, $N_{c}$ is the number of instances belonging to the $c^{th}$ prototype $\boldsymbol{p}^{c}_{(i)}$ in $\mathbb{V}_{(i)}$, and $\tau$ represents the temperature for contrastive learning.
\hc{It is worth noting that} the na\"ive prototype contrastive learning (denoted as NPC) using original sequences is a special case of the proposed MPC scheme when $q=1$ and $x=0$ (see \hc{Eq. (\ref{eq_2}) and (\ref{eq_3})}).
The MPC scheme can be viewed as to perform finer prototype learning with different subsequences, and allow the model to jointly attend to key skeleton patterns from different representation subspaces of the original sequences, which encourages capturing more discriminative skeleton features for person re-ID (see Sec. \ref{further}).
The objective of MPC can be theoretically formulated in the form of Expectation-Maximization (EM) algorithms. We prove the effectiveness of MPC and show its relations to existing contrastive losses in \hc{Appendix A}.

\subsection{Masked Intra-Sequence Contrastive Learning}
\label{MIC_sec}
The continuity of human motion typically results in very little variation of poses/skeletons within a small temporal interval \cite{rao2021self}. Due to this nature, subsequences of the same skeleton sequence usually possess strong inherent correlations. For example, they could locally share similar skeletons and partial sequences with consistent walking patterns.
To exploit such intra-sequence relationships and inherent consistency ($e.g.,$ pattern invariance) within sequences to learn better skeleton representations, we propose the \textbf{\textit{masked intra-sequence contrastive learning (MIC)}} below.

Given two skeleton instances ($i.e.,$ subsequence representations), $\boldsymbol{v}_{(i)}$ and $\boldsymbol{v}_{(j)}$, of the same sequence, we first map them into a contrasting space $\mathbb{R}^{H}$ with a fully-connected (FC) layer $\mathcal{F}_{c}(\cdot)$ by: $\mathcal{F}_{c}\left(\boldsymbol{v}_{(i)}\right)\!=\!\boldsymbol{z}_{(i)}$ \hc{and} $\mathcal{F}_{c}\left(\boldsymbol{v}_{(j)}\right)\!=\!\boldsymbol{z}_{(j)}$, where $\boldsymbol{z}_{(i)}, \boldsymbol{z}_{(j)} \in \mathbb{R}^{H}$. 
Inspired by \cite{chen2021exploring}, we leverage a Siamese architecture to contrast one instance in the original feature space with the other one in the new contrasting space, so as to \textit{symmetrically} learn their inherent similarity. To this end, we exploit a masked intra-sequence contrastive learning (MIC) loss to minimize the negative cosine similarity between two instances of the same sequence by:
\begin{equation}
\mathcal{L}_{\text {MIC}}=-\alpha\frac{\boldsymbol{z}_{(i)}}{\left\|\boldsymbol{z}_{(i)}\right\|_{2}} \cdot \frac{\boldsymbol{v}_{(j)}}{\left\|\boldsymbol{v}_{(j)}\right\|_{2}}-\beta\frac{\boldsymbol{z}_{(j)}}{\left\|\boldsymbol{z}_{(j)}\right\|_{2}} \cdot \frac{\boldsymbol{v}_{(i)}}{\left\|\boldsymbol{v}_{(i)}\right\|_{2}},
\label{eq_6}
\end{equation}
where $\|\cdot\|_{2}$ denotes $\ell_{2}$-norm, $\alpha$ and $\beta$ are weights for contrastive learning of representation pairs $(\boldsymbol{z}_{(i)},\boldsymbol{v}_{(j)})$ and $(\boldsymbol{z}_{(j)},\boldsymbol{v}_{(i)})$, respectively. Here $\mathcal{L}_{\text {MIC}}$ is defined for two subsequence representations of a skeleton sequence and the total loss is averaged over all sequences.
To enable more stable and better contrastive learning, we employ a symmetrized loss with equal weights for two contrastive representation pairs, $i.e.$, $\alpha=\beta=0.5$, and adopt an alternating stop-gradient operation following \cite{chen2021exploring} when contrasting each pair, as shown in Fig. \ref{model} (Note that we only visualize one contrastive pair for conciseness). We provide hypotheses and proof for the effectiveness of MIC in \hc{Appendix A}.

\begin{table*}[t]
\centering
\scalebox{0.66}{
\setlength{\tabcolsep}{2.2mm}{
\begin{tabular}{llrrrrrrrrrrrrrr}
\hline
\textbf{}                                                                                         & \textbf{}                                         & \multicolumn{1}{l}{\textbf{}}          & \multicolumn{1}{l}{\textbf{}}       & \multicolumn{4}{c}{\textbf{KS20}}                                                                                                                & \multicolumn{4}{c}{\textbf{KGBD}}                                                                                                                & \multicolumn{4}{c}{\textbf{IAS-A}}                                                                                                               \\ \hline
\textbf{Types}                                                                                    & \textbf{Methods}                                  & \multicolumn{1}{c}{\textbf{\# Params}} & \multicolumn{1}{c}{\textbf{GFLOPs}} & \multicolumn{1}{c}{\textbf{top-1}} & \multicolumn{1}{c}{\textbf{top-5}} & \multicolumn{1}{c}{\textbf{top-10}} & \multicolumn{1}{c}{\textbf{mAP}} & \multicolumn{1}{c}{\textbf{top-1}} & \multicolumn{1}{c}{\textbf{top-5}} & \multicolumn{1}{c}{\textbf{top-10}} & \multicolumn{1}{c}{\textbf{mAP}} & \multicolumn{1}{c}{\textbf{top-1}} & \multicolumn{1}{c}{\textbf{top-5}} & \multicolumn{1}{c}{\textbf{top-10}} & \multicolumn{1}{c}{\textbf{mAP}} \\ \hline
\multirow{2}{*}{\textbf{Hand-crafted}}                                                            & $D_{13}$ \cite{munaro2014one}    & —                                      & —                                   & 39.4                               & 71.7                               & 81.7                                & 18.9                             & 17.0                               & 34.4                               & 44.2                                & 1.9                              & 40.0                               & 58.7                               & 67.6                                & 24.5                             \\
                                                                                                  & $D_{16}$ \cite{pala2019enhanced} & —                                      & —                                   & 51.7                               & 77.1                               & 86.9                                & 24.0                             & 31.2                               & 50.9                               & 59.8                                & 4.0                              & 42.7                               & 62.9                               & 70.7                                & 25.2                             \\ \hline
\multirow{4}{*}{\textbf{Supervised}}                                                              & PoseGait \cite{liao2020model}    & 8.93M                                  & 121.60                              & 49.4                               & 80.9                               & 90.2                                & 23.5                             & 50.6                               & 67.0                               & 72.6                                & 13.9                             & 28.4                               & 55.7                               & 69.2                                & 17.5                             \\
                                                                                                  & SGELA \cite{rao2021self} + DF    & 9.09M                                  & 7.48                                & 49.7                               & 67.0                               & 77.1                                & 22.2                             & 43.7                               & 58.7                               & 65.0                                & 7.1                              & 18.0                               & 32.1                               & 46.2                                & 13.5                             \\
                                                                                                  & MG-SCR \cite{rao2021multi}       & 0.35M                                  & 6.60                                & 46.3                               & 75.4                               & 84.0                                & 10.4                             & 44.0                               & 58.7                               & 64.6                                & 6.9                              & 36.4                               & 59.6                               & 69.5                                & 14.1                             \\
                                                                                                  & SM-SGE \cite{rao2021sm} + DF     & 6.25M                                  & 23.92                               & 49.8                               & 78.1                               & 85.2                                & 11.7                             & 43.2                               & 58.6                               & 64.6                                & 7.5                              & 38.5                               & 63.2                               & 73.9                                & 15.0                             \\ \hline
\multirow{4}{*}{\textbf{\begin{tabular}[c]{@{}l@{}}Self-supervised\\ /Unsupervised\end{tabular}}} & AGE \cite{rao2020self}           & 7.15M                                  & 37.37                               & 43.2                               & 70.1                               & 80.0                                & 8.9                              & 2.9                                & 5.6                                & 7.5                                 & 0.9                              & 31.1                               & 54.8                               & 67.4                                & 13.4                             \\
                                                                                                  & SGELA \cite{rao2021self}         & 8.47M                                  & 7.47                                & 45.0                               & 65.0                               & 75.1                                & 21.2                             & 38.1                               & 53.5                               & 60.0                                & 4.5                              & 16.7                               & 30.2                               & 44.0                                & 13.2                             \\
                                                                                                  & SM-SGE \cite{rao2021sm}          & 5.58M                                  & 22.61                               & 45.9                               & 71.9                               & 81.2                                & 9.5                              & 38.2                               & 54.2                               & 60.7                                & 4.4                              & 34.0                               & 60.5                               & 71.6                                & 13.6                             \\
                                                                                                  & \textbf{SimMC (Ours)}                                      & 0.15M                                  & 0.99                                & \textbf{66.4}                      & \textbf{80.7}                      & \textbf{87.0}                       & \textbf{22.3}                    & \textbf{54.9}                      & \textbf{66.2}                      & \textbf{70.6}                       & \textbf{11.7}                    & \textbf{44.8}                      & \textbf{65.3}                      & \textbf{72.9}                       & \textbf{18.7}                    \\ \hline
\multirow{3}{*}{\textbf{\begin{tabular}[c]{@{}l@{}}Unsupervised \\ Fine-tuinig\end{tabular}}}     & SGELA + SimMC                                     & 8.80M                                  & 10.10                               & \textit{47.3}                      & \textit{69.7}                      & \textit{79.3}                       & 20.1                             & \textit{51.7}                      & \textit{62.7}                      & \textit{67.9}                       & \textit{15.1}                    & \textit{16.8}                      & \textit{33.3}                      & \textit{48.7}                       & 12.0                    \\
                                                                                                  & MG-SCR + SimMC                                    & 0.53M                                  & 7.88                                & \textit{71.1}                      & \textit{83.6}                      & \textit{89.1}                       & \textit{22.7}                    & \textit{47.4}                      & \textit{59.3}                      & \textit{64.9}                       & \textit{11.0}                    & \textit{47.2}                      & \textit{69.0}                      & \textit{77.3}                       & \textit{22.4}                    \\
                                                                                                  & SM-SGE + SimMC                                    & 5.89M                                  & 25.10                               & \textit{67.2}                      & \textit{82.2}                      & \textit{88.5}                       & \textit{23.0}                    & \textit{47.1}                      & \textit{59.2}                      & \textit{64.9}                       & \textit{10.8}                    & \textit{51.3}                      & \textit{69.9}                      & \textit{75.6}                       & \textit{27.3}                    \\ \hline
\end{tabular}
}
}
\caption{Performance comparison with existing state-of-the-art skeleton-based methods on KS20, KGBD, and IAS-A. The amount of network parameters (million (M)) and computational complexity (giga floating-point operations (GFLOPs)) for the deep learning based methods are reported. ``+ DF'' denotes direct supervised fine-tuning. \textbf{Bold} refers to the best cases among self-supervised/unsupervised methods, while \textit{italics} indicate achieving higher performance when exploiting SimMC (``+ SimMC'') to fine-tune corresponding pre-trained representations. }
\label{KS20_KGBD_BIWIS}
\end{table*}

\begin{table*}[t]
\centering
\scalebox{0.66}{
\setlength{\tabcolsep}{3.5mm}{
\begin{tabular}{llrrrrrrrrrrrr}
\hline
\textbf{}                                                                                         & \textbf{}                                         & \multicolumn{4}{c}{\textbf{IAS-B}}                                                                                                               & \multicolumn{4}{c}{\textbf{BIWI-W}}                                                                                                              & \multicolumn{4}{c}{\textbf{BIWI-S}}                                                                                                              \\ \hline
\textbf{Types}                                                                                    & \textbf{Methods}                                  & \multicolumn{1}{c}{\textbf{top-1}} & \multicolumn{1}{c}{\textbf{top-5}} & \multicolumn{1}{c}{\textbf{top-10}} & \multicolumn{1}{c}{\textbf{mAP}} & \multicolumn{1}{c}{\textbf{top-1}} & \multicolumn{1}{c}{\textbf{top-5}} & \multicolumn{1}{c}{\textbf{top-10}} & \multicolumn{1}{c}{\textbf{mAP}} & \multicolumn{1}{c}{\textbf{top-1}} & \multicolumn{1}{c}{\textbf{top-5}} & \multicolumn{1}{c}{\textbf{top-10}} & \multicolumn{1}{c}{\textbf{mAP}} \\ \hline
\multirow{2}{*}{\textbf{Hand-crafted}}                                                            & $D_{13}$ \cite{munaro2014one}    & 43.7                               & 68.6                               & 76.7                                & 23.7                             & 14.2                               & 20.6                               & 23.7                                & 17.2                             & 28.3                               & 53.1                               & 65.9                                & 13.1                             \\
                                                                                                  & $D_{16}$ \cite{pala2019enhanced} & 44.5                               & 69.1                               & 80.2                                & 24.5                             & 17.0                               & 25.3                               & 29.6                                & 18.8                             & 32.6                               & 55.7                               & 68.3                                & 16.7                             \\ \hline
\multirow{4}{*}{\textbf{Supervised}}                                                              & PoseGait \cite{liao2020model}    & 28.9                               & 51.6                               & 62.9                                & 20.8                             & 8.8                                & 23.0                               & 31.2                                & 11.1                             & 14.0                               & 40.7                               & 56.7                                & 9.9                              \\
                                                                                                  & SGELA \cite{rao2021self} + DF    & 23.6                               & 42.9                               & 51.9                                & 14.8                             & 13.9                               & 15.3                               & 16.7                                & 22.9                             & 29.2                               & 65.2                               & 73.8                                & 23.5                             \\
                                                                                                  & MG-SCR \cite{rao2021multi}       & 32.4                               & 56.5                               & 69.4                                & 12.9                             & 10.8                               & 20.3                               & 29.4                                & 11.9                             & 20.1                               & 46.9                               & 64.1                                & 7.6                              \\
                                                                                                  & SM-SGE \cite{rao2021sm} + DF     & 44.3                               & 68.2                               & 77.5                                & 14.9                             & 16.7                               & 31.0                               & 40.2                                & 18.7                             & 34.8                               & 60.6                               & 71.5                                & 12.8                             \\ \hline
\multirow{4}{*}{\textbf{\begin{tabular}[c]{@{}l@{}}Self-supervised\\ /Unsupervised\end{tabular}}} & AGE \cite{rao2020self}           & 31.1                               & 52.3                               & 64.2                                & 12.8                             & 11.7                               & 21.4                               & 27.3                                & 12.6                             & 25.1                               & 43.1                               & 61.6                                & 8.9                              \\
                                                                                                  & SGELA \cite{rao2021self}         & 22.2                               & 40.8                               & 50.2                                & 14.0                             & 11.7                               & 14.0                               & 14.7                                & 19.0                             & 25.8                               & 51.8                               & 64.4                                & \textbf{15.1}                    \\
                                                                                                  & SM-SGE \cite{rao2021sm}          & 38.9                               & 64.1                               & 75.8                                & 13.3                             & 13.2                               & 25.8                               & 33.5                                & 15.2                             & 31.3                               & 56.3                               & 69.1                                & 10.1                             \\
                                                                                                  & SimMC (Ours)                             & \textbf{46.3}                      & \textbf{68.1}                      & \textbf{77.0}                       & \textbf{22.9}                    & \textbf{24.5}                      & \textbf{36.7}                      & \textbf{44.5}                       & \textbf{19.9}                    & \textbf{41.7}                      & \textbf{66.6}                      & \textbf{76.8}                       & 12.3                             \\ \hline
\multirow{3}{*}{\textbf{\begin{tabular}[c]{@{}l@{}}Unsupervised \\ Fine-tuinig\end{tabular}}}     & SGELA + SimMC                                     & 21.2                               & 39.1                               & 48.8                                & 14.0                             & \textit{18.4}                      & \textit{23.1}                      & \textit{25.0}                       & \textit{28.7}                    & \textit{51.8}                      & \textit{71.3}                      & \textit{74.4}                       & \textit{43.3}                    \\
                                                                                                  & MG-SCR + SimMC                                    & \textit{52.4}                      & \textit{72.0}                      & \textit{78.8}                       & \textit{29.1}                    & \textit{25.1}                      & \textit{37.5}                      & \textit{46.4}                       & \textit{20.3}                    & \textit{28.3}                      & \textit{51.6}                      & \textit{64.8}                       & \textit{10.9}                    \\
                                                                                                  & SM-SGE + SimMC                                    & \textit{55.3}                      & \textit{72.6}                      & \textit{80.3}                       & \textit{34.1}                    & \textit{25.9}                      & \textit{39.2}                      & \textit{45.2}                       & \textit{22.4}                    & \textit{42.6}                      & \textit{64.8}                      & \textit{76.2}                       & \textit{15.4}                    \\ \hline
\end{tabular}
}
}
\caption{Performance comparison on IAS-B, BIWI-Walking (BIWI-W), and BIWI-Still (BIWI-S). 
\textbf{Bold} refers to the best cases among self-supervised/unsupervised methods, while \textit{italics} indicate achieving higher performance with the fine-tuning of SimMC.}
\label{IASA_IASB_BIWIW}
\end{table*}
\subsection{The Entire Framework}
The proposed SimMC framework combines both MPC loss (see Eq. (\ref{eq_5})) and MIC loss (see Eq. (\ref{eq_6})) to perform unsupervised contrastive learning of skeleton representations with:
\begin{equation}
\mathcal{L}=\lambda\mathcal{L}_{\text {MIC}}+(1-\lambda)\mathcal{L}_{\text {MPC}},
\label{eq_7}
\end{equation}
where $\lambda$ is the weight coefficient to trade off the importance of different contrastive learning. For convenience, here we use $\mathcal{L}_{\text {MIC}}$ to denote the total MIC loss averaging over all training skeleton sequences. 
To facilitate training and generate more reliable clusters, we optimize our model by alternating clustering and contrastive representation learning.
For the person re-ID task, we exploit the encoder $\psi(\cdot)$ learned by our framework to encode each skeleton sequence of the probe set $\Phi_{\mathcal{P}}$ into corresponding representations, $\{\boldsymbol{v}^{\mathcal{P}}_{i}\}_{i=1}^{N_{2}}$, which are matched with the representations, $\{\boldsymbol{v}^{\mathcal{G}}_{j}\}_{j=1}^{N_{3}}$, of the same identity in the gallery set $\Phi_{\mathcal{G}}$ based on the Euclidean distance.

\section{Experiments}
\subsection{Experimental Settings}
\paragraph{Datasets.} We evaluate our framework on four person re-ID benchmark datasets with 3D skeleton data, namely \textit{IAS-Lab} \cite{munaro2014feature}, \textit{KS20} \cite{nambiar2017context}, \textit{BIWI} \cite{munaro2014one}, \textit{KGBD} \cite{andersson2015person}, and a large-scale multi-view gait dataset \textit{CASIA-B} \cite{yu2006framework}, which contain 11, 20, 50, 164, and 124 different individuals, respectively. For BIWI and IAS-Lab, we set each testing set as the gallery and the other one as the probe. For KS20, we randomly take one skeleton sequence from each view as the probe sequence and use one half of the remaining sequences for training and the other half as the gallery. For KGBD, we randomly choose one skeleton video of each individual as the probe set, and equally divide the remaining videos into the training set and gallery set. In CASIA-B, all testing sequences are grouped by three conditions (Normal, Bags, Clothes), and we evaluate our framework with single-condition and cross-condition settings following \cite{liu2015enhancing}. We repeat experiments with each setup for multiple times and report the average performance.

\paragraph{Implementation Details.} We set sequence length $f$ to $6$ on IAS-Lab, KS20, BIWI, and KGBD datasets for a fair comparison with existing methods, \hc{and} empirically employ $x=2$ random masks for subsequence sampling. For the largest dataset CASIA-B with roughly estimated skeleton data from RGB videos, we set $f=40$ with $x=10$ random masks. The number of random subsequence sampling is $q=2$ and the embedding size for skeleton representations is $H=256$ for all datasets. 
We empirically set the temperature $\tau=0.06$ (KGBD), $\tau=0.07$ (BIWI), $\tau=0.075$ (CASIA-B), $\tau=0.08$ (KS20, IAS-Lab) for MPC learning, and adopt the weight coefficient $\lambda=0.5$ for KS20, KGBD, and IAS-B, $\lambda=0.75$ for IAS-A, and $\lambda=0.25$ for BIWI and CASIA-B.
We employ Adam optimizer with learning rate $0.00035$ and batch size $256$ for all datasets. To perform unsupervised fine-tuning with SimMC, we train SimMC on the unlabeled skeleton representations pre-trained by original models, and exploit the skeleton representations learned by SimMC for person re-ID. More implementation details are provided in \hc{Appendix B}.

\paragraph{Evaluation Metrics.}
We compute Cumulative Matching Characteristics (CMC) curve and adopt top-1/top-5/top-10 accuracy and Mean Average Precision (mAP) \cite{zheng2015scalable} to quantitatively evaluate person re-ID performance.

\subsection{Comparison with State-of-the-Arts}
\label{comparison}
We compare our framework with existing state-of-the-art self-supervised and unsupervised skeleton-based methods on KS20, KGBD, IAS-Lab, and BIWI in Table \ref{KS20_KGBD_BIWIS} and \ref{IASA_IASB_BIWIW}.
We also include the latest supervised skeleton-based methods and representative hand-crafted methods as a performance reference. 


\begin{table*}[t]
\centering
\scalebox{0.68}{
\setlength{\tabcolsep}{1.18mm}{
\begin{tabular}{lrrrrrrrrrrrrrrrrrrrr}
\hline
\textbf{Probe-Gallery}                            & \multicolumn{4}{c}{\textbf{Normal-Normal}}                                                                                                                 & \multicolumn{4}{c}{\textbf{Bags-Bags}}                                                                                                                 & \multicolumn{4}{c}{\textbf{Clothes-Clothes}}                                                                                                                 & \multicolumn{4}{c}{\textbf{Clothes-Normal}}                                                                                                                 & \multicolumn{4}{c}{\textbf{Bags-Normal}}                                                                                                                 \\ \hline
\textbf{Methods}                                  & \multicolumn{1}{c}{\textbf{top-1}} & \multicolumn{1}{c}{\textbf{top-5}} & \multicolumn{1}{c}{\textbf{top-10}} & \multicolumn{1}{c}{\textbf{mAP}} & \multicolumn{1}{c}{\textbf{top-1}} & \multicolumn{1}{c}{\textbf{top-5}} & \multicolumn{1}{c}{\textbf{top-10}} & \multicolumn{1}{c}{\textbf{mAP}} & \multicolumn{1}{c}{\textbf{top-1}} & \multicolumn{1}{c}{\textbf{top-5}} & \multicolumn{1}{c}{\textbf{top-10}} & \multicolumn{1}{c}{\textbf{mAP}} & \multicolumn{1}{c}{\textbf{top-1}} & \multicolumn{1}{c}{\textbf{top-5}} & \multicolumn{1}{c}{\textbf{top-10}} & \multicolumn{1}{c}{\textbf{mAP}} & \multicolumn{1}{c}{\textbf{top-1}} & \multicolumn{1}{c}{\textbf{top-5}} & \multicolumn{1}{c}{\textbf{top-10}} & \multicolumn{1}{c}{\textbf{mAP}} \\ \hline
ELF \cite{gray2008viewpoint}     & 12.3                               & 35.6                               & 50.3                                & —                                & 5.8                                & 25.5                               & 37.6                                & —                                & 19.9                               & 43.9                               & 56.7                                & —                                & 5.6                                & 16.0                               & 26.3                                & —                                & 17.1                               & 30.0                               & 37.9                                & —                                \\
SDALF \cite{farenzena2010person} & 4.9                                & 27.0                               & 41.6                                & —                                & 10.2                               & 33.5                               & 47.2                                & —                                & 16.7                               & 42.0                               & 56.7                                & —                                & 11.6                               & 19.4                               & 27.6                                & —                                & 22.9                               & 30.1                               & 36.1                                & —                                \\
MLR \cite{liu2015enhancing}      & 16.3                               & 43.4                               & 60.8                                & —                                & 18.9                               & 44.8                               & 59.4                                & —                                & 25.4                               & 53.3                               & 68.9                                & —                                & 20.3                               & 42.6                               & \textbf{56.9}                       & —                                & 31.8                               & 53.6                               & 64.1                                & —                                \\
AGE \cite{rao2020self}           & 20.8                               & 29.3                               & 34.2                                & 3.5                              & 37.1                               & 56.2                               & 67.0                                & 9.8                              & 35.5                               & 54.3                               & 65.3                                & 9.6                              & 14.6                               & 33.0                               & 42.7                                & 3.0                              & 32.4                               & 51.2                               & 60.1                                & 3.9                              \\
SM-SGE \cite{rao2021sm}          & 50.2                               & 73.5                               & 81.9                                & 6.6                              & 26.6                               & 49.0                               & 59.4                                & 9.3                              & 27.2                               & 51.4                               & 63.2                                & 9.7                              & 10.6                               & 26.3                               & 35.9                                & 3.0                              & 16.6                               & 36.8                               & 47.5                                & 3.5                              \\
SGELA \cite{rao2021self}         & 71.8                               & 87.5                               & 91.4                                & 9.8                              & 48.1                               & 69.5                               & 77.7                                & \textbf{16.5}                    & 51.2                               & 73.8                               & 81.5                                & 7.1                              & 15.9                               & 30.8                               & 40.6                                & 4.7                              & 36.4                               & 57.1                               & 64.6                                & 6.7                              \\
SimMC (Ours)                                      & \textbf{84.8}                      & \textbf{92.3}                      & \textbf{93.7}                       & \textbf{10.8}                    & \textbf{69.1}                      & \textbf{86.6}                      & \textbf{91.3}                       & \textbf{16.5}                    & \textbf{68.0}                      & \textbf{88.1}                      & \textbf{93.0}                       & \textbf{15.7}                    & \textbf{25.6}                      & \textbf{43.8}                      & 54.0                                & \textbf{5.4}                     & \textbf{42.0}                      & \textbf{59.8}                      & \textbf{68.9}                       & \textbf{7.1}                     \\ \hline
\end{tabular}
}
}
\caption{Comparison with appearance-based and skeleton-based methods on CASIA-B. ``Bags-Normal'' represents the probe set with ``Bags'' condition and gallery set with ``Normal'' condition. ``—'' indicates no published result. Full results are in \hc{Appendix B}. }
\label{CASIA-B}
\end{table*}

\begin{table*}[t]
\centering
\scalebox{0.68}{
\setlength{\tabcolsep}{5.5mm}{
\begin{tabular}{lrrrrrrrrrrrr}
\hline
\textbf{}               & \multicolumn{2}{c}{\textbf{IAS-A}}                                    & \multicolumn{2}{c}{\textbf{IAS-B}}                                    & \multicolumn{2}{c}{\textbf{BIWI-S}}                                   & \multicolumn{2}{c}{\textbf{BIWI-W}}                                   & \multicolumn{2}{c}{\textbf{KS20}}                                     & \multicolumn{2}{c}{\textbf{KGBD}}                                     \\ \hline
\textbf{Configurations} & \multicolumn{1}{c}{\textbf{top-1}} & \multicolumn{1}{c}{\textbf{mAP}} & \multicolumn{1}{c}{\textbf{top-1}} & \multicolumn{1}{c}{\textbf{mAP}} & \multicolumn{1}{c}{\textbf{top-1}} & \multicolumn{1}{c}{\textbf{mAP}} & \multicolumn{1}{c}{\textbf{top-1}} & \multicolumn{1}{c}{\textbf{mAP}} & \multicolumn{1}{c}{\textbf{top-1}} & \multicolumn{1}{c}{\textbf{mAP}} & \multicolumn{1}{c}{\textbf{top-1}} & \multicolumn{1}{c}{\textbf{mAP}} \\ \hline
\textbf{Baseline}       & 29.4                               & 13.8                             & 30.2                               & 13.3                             & 24.8                               & 9.3                              & 10.9                               & 14.1                             & 17.0                               & 9.5                              & 34.5                               & 6.4                              \\
\textbf{NPC}            & 39.2                               & 17.8                             & 40.7                               & 21.5                             & 38.1                               & 11.3                             & 21.2                               & 18.3                             & 64.8                               & 20.5                             & 53.0                               & 11.0                             \\
\textbf{MPC}            & 43.1                               & 18.5                             & 43.8                               & 22.3                             & 40.1                               & 11.7                             & 23.7                               & 19.5                             & 65.6                               & 21.1                             & 53.6                               & 11.0                             \\
\textbf{MPC + MIC}      & 44.8                               & 18.7                             & 46.3                               & 22.9                             & 41.7                               & 12.3                             & 24.5                               & 19.9                             & 66.4                               & 22.3                             & 54.9                               & 11.7                             \\ \hline
\end{tabular}
}
}
\caption{Ablation study of framework with different configurations: Na\"ive prototype contrastive learning (NPC) using only original sequences, masked prototype contrastive learning (MPC) scheme and corresponding masked intra-sequence contrastive learning (MIC). }
\label{ablation}
\end{table*}

\paragraph{Comparison with Self-supervised and Unsupervised Methods.}
Our framework shows evident advantages in terms of performance and efficiency over existing state-of-the-art self-supervised and unsupervised methods. As reported in Table \ref{KS20_KGBD_BIWIS} and \ref{IASA_IASB_BIWIW}, SimMC significantly outperforms AGE \cite{rao2020self} and SM-SGE \cite{rao2021sm} that manually design pretext tasks based on pre-defined skeleton modeling such as skeleton graphs by a large margin of $7.4$-$52.0\%$ top-1 accuracy and $2.2$-$13.4\%$ mAP on all datasets. Compared with the SGELA model \cite{rao2021self} using direct inter-sequence contrastive learning, our framework achieves remarkably better performance on five out of six testing sets (KS20, KGBD, IAS-A, IAS-B, BIWI-W) by up to $28.1\%$ top-1 accuracy and $8.9\%$ mAP, which demonstrates that the proposed SimMC combining both prototype (MPC) and intra-sequence contrastive learning (MIC) can capture more discriminative features within skeleton sequences for person re-ID on different datasets. 
Notably, SimMC also enjoys the smallest model size (only 0.15M) for skeleton representation learning among all approaches shown in Table \ref{KS20_KGBD_BIWIS}, which suggests its higher model efficiency for person re-ID tasks.

By applying the proposed framework to fine-tuning SGELA and SM-SGE models, we can further improve their performance with an average gain of $16.9\%$ and $8.1\%$ top-1 accuracy respectively on all datasets. Such results demonstrate both effectiveness and scalability of proposed masked contrastive learning, which is compatible with existing models and can fully exploit their pre-trained features to achieve higher-quality skeleton representations for person re-ID.

\paragraph{Comparison with Hand-crafted and Supervised Methods.}
In contrast to hand-crafted methods ($D_{13}$ and $D_{16}$) that rely on geometric joint distances and anthropometric descriptors, our approach obtains similar performance on IAS testing sets, while it achieves a distinct improvement of $7.5$-$37.9\%$ top-1 accuracy on BIWI, KS20, and KGBD datasets that contain more views and individuals. 
Despite utilizing \textit{unlabeled} skeleton data as the sole input, the proposed SimMC still performs better than the latest supervised models PoseGait and MG-SCR in most cases. 
Interestingly, applying SimMC to SM-SGE achieves significantly higher performance gains than direct supervised fine-tuing (DF) in terms of top-1 accuracy ($3.9$-$17.4\%$), top-5 accuracy ($0.6$-$6.7\%$), top-10 accuracy ($0.3$-$5.0\%$), and mAP ($3.3$-$19.2\%$) on all datasets.
With highly efficient performance and strong scalability, the proposed unsupervised SimMC can be a more general framework for skeleton-based person re-ID and related tasks. 

\begin{figure}[t]
    \centering
      \subfigure[AGE]{\scalebox{0.21}{\label{AGE_1_10_BIWI}\includegraphics[]{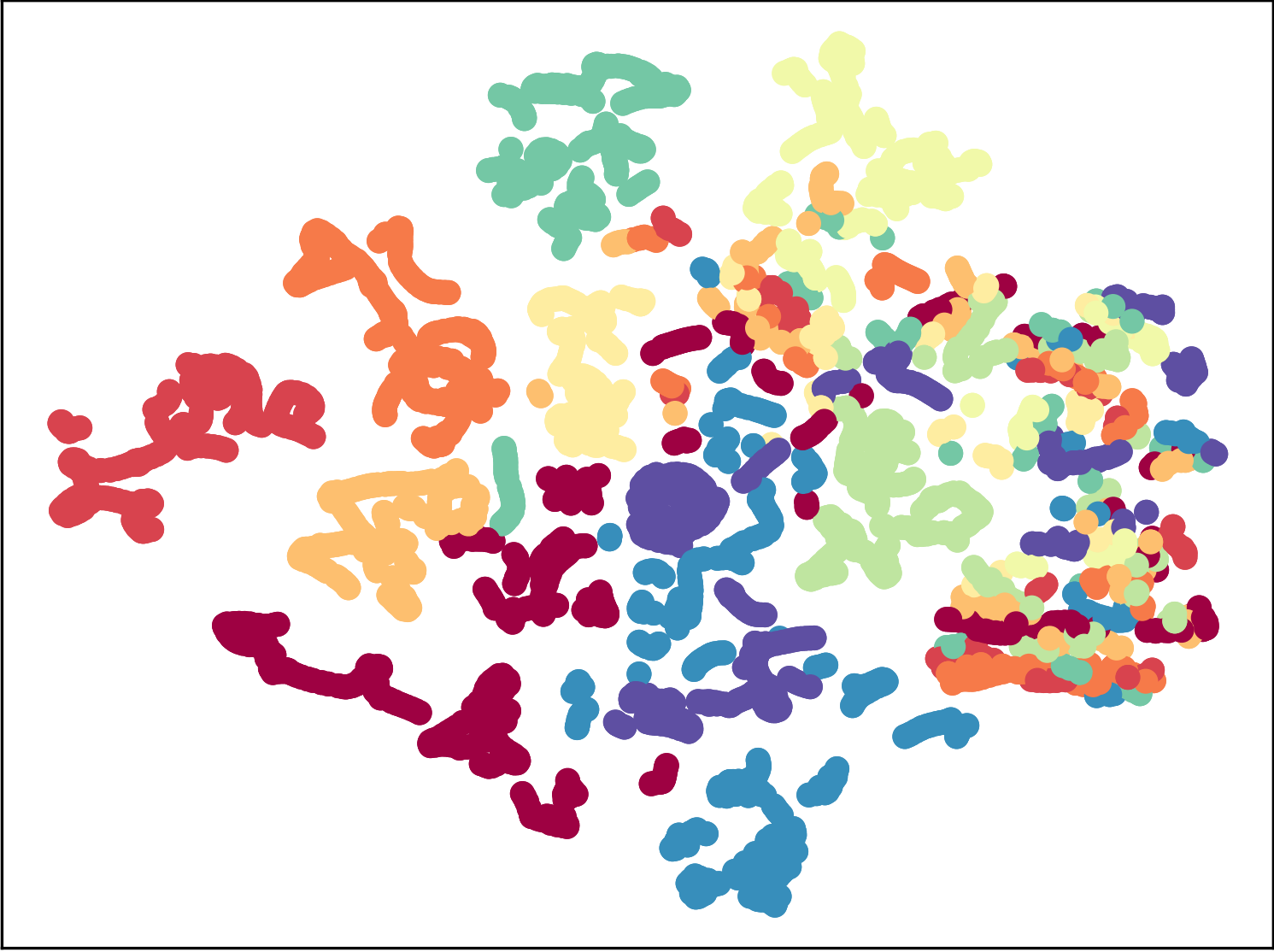}}}  \ \ 
      \subfigure[SM-SGE]{\scalebox{0.21}{\label{SMSGE_1_10_BIWI}\includegraphics[]{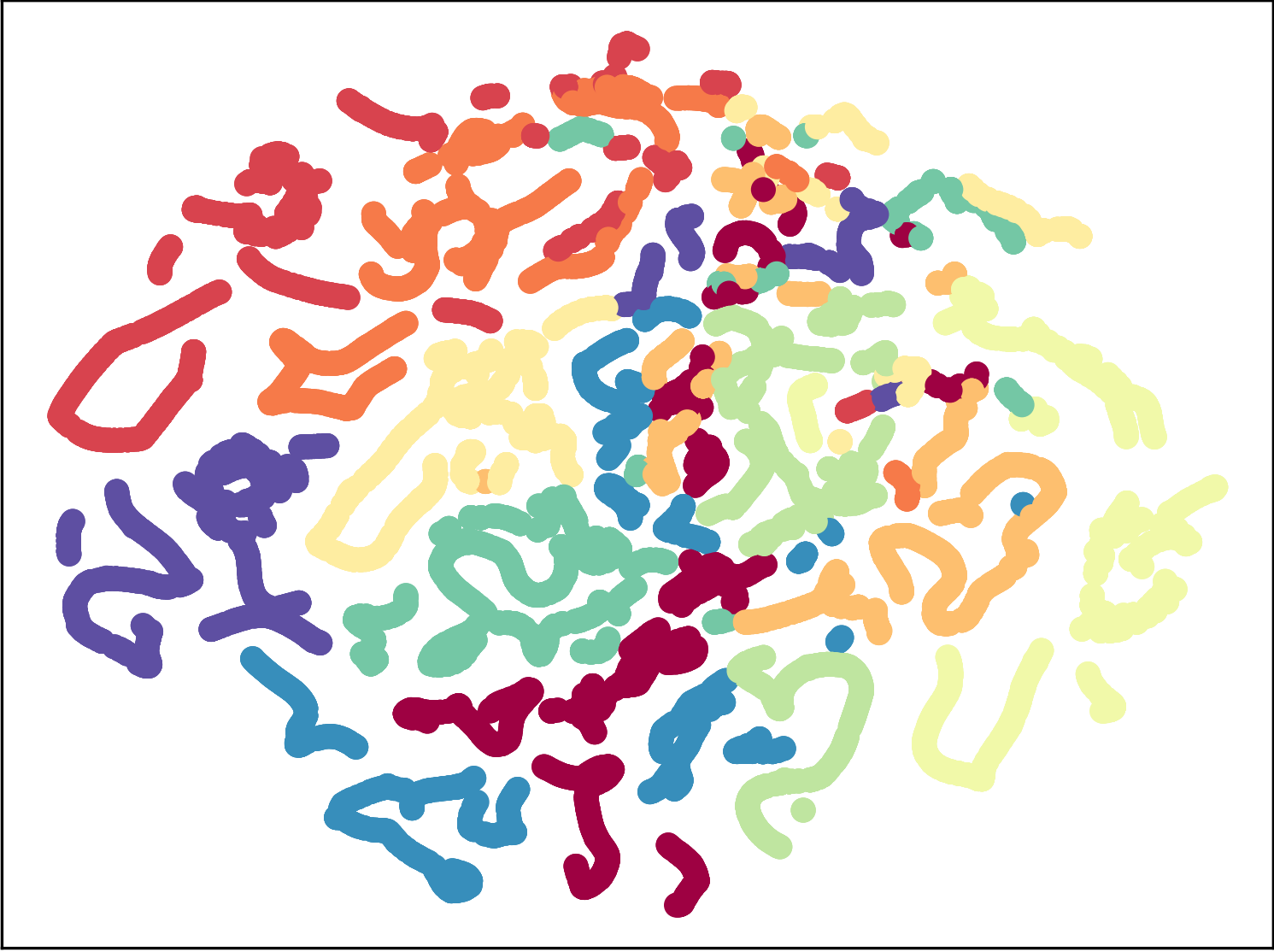}}}  
        \ \
      \subfigure[SimMC]{\scalebox{0.21}{\label{SPC_1_10_BIWI}\includegraphics[]{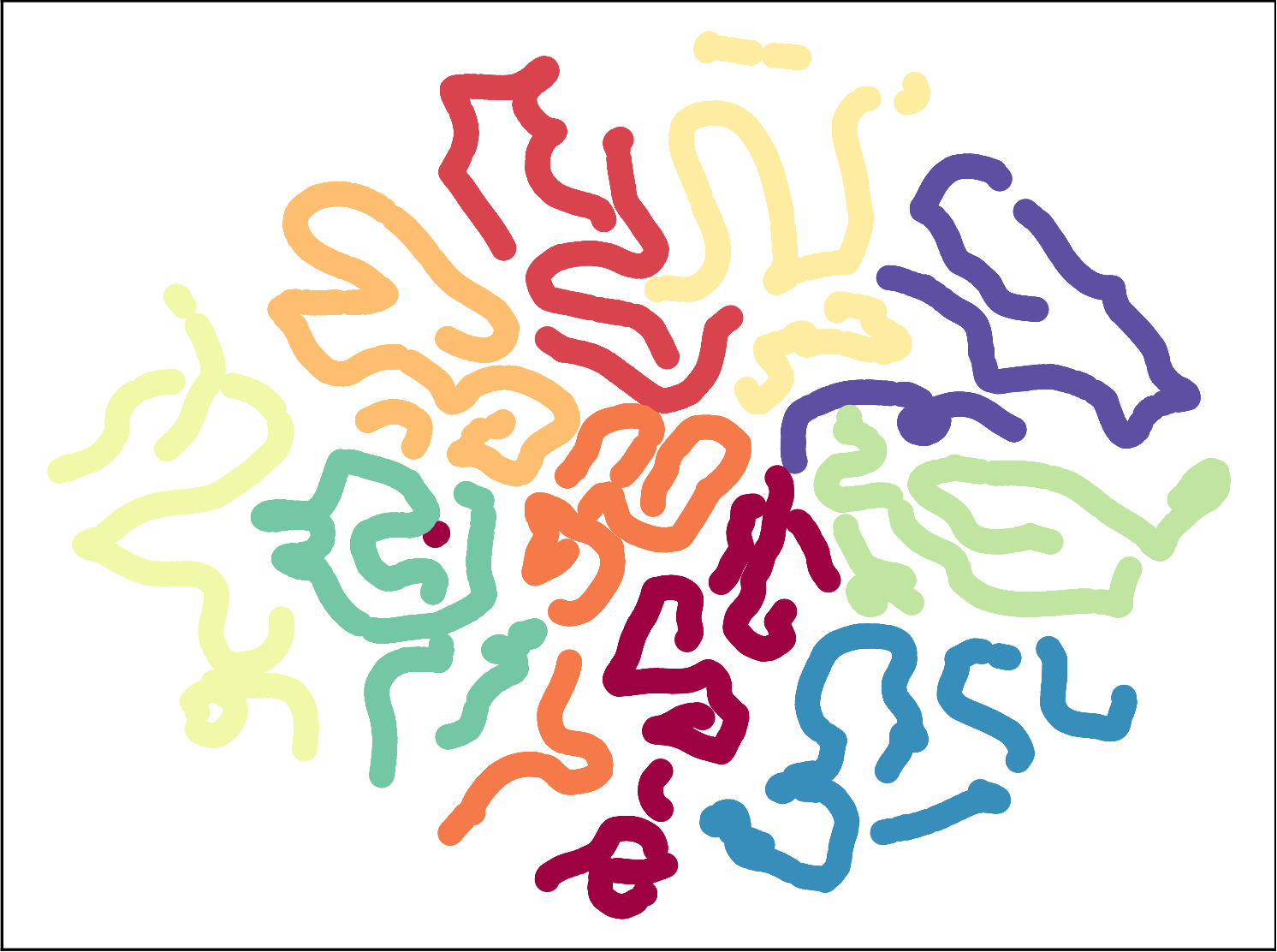}}}
    \caption{t-SNE visualization of representations learned by AGE (a), SM-SGE (b), and SimMC (c) for first ten classes in BIWI. Different colors denote skeleton representations of different classes.}
    \label{TNS_comp}
\end{figure}

 \begin{figure}[t]
    \centering
    \scalebox{0.25}{\includegraphics[]{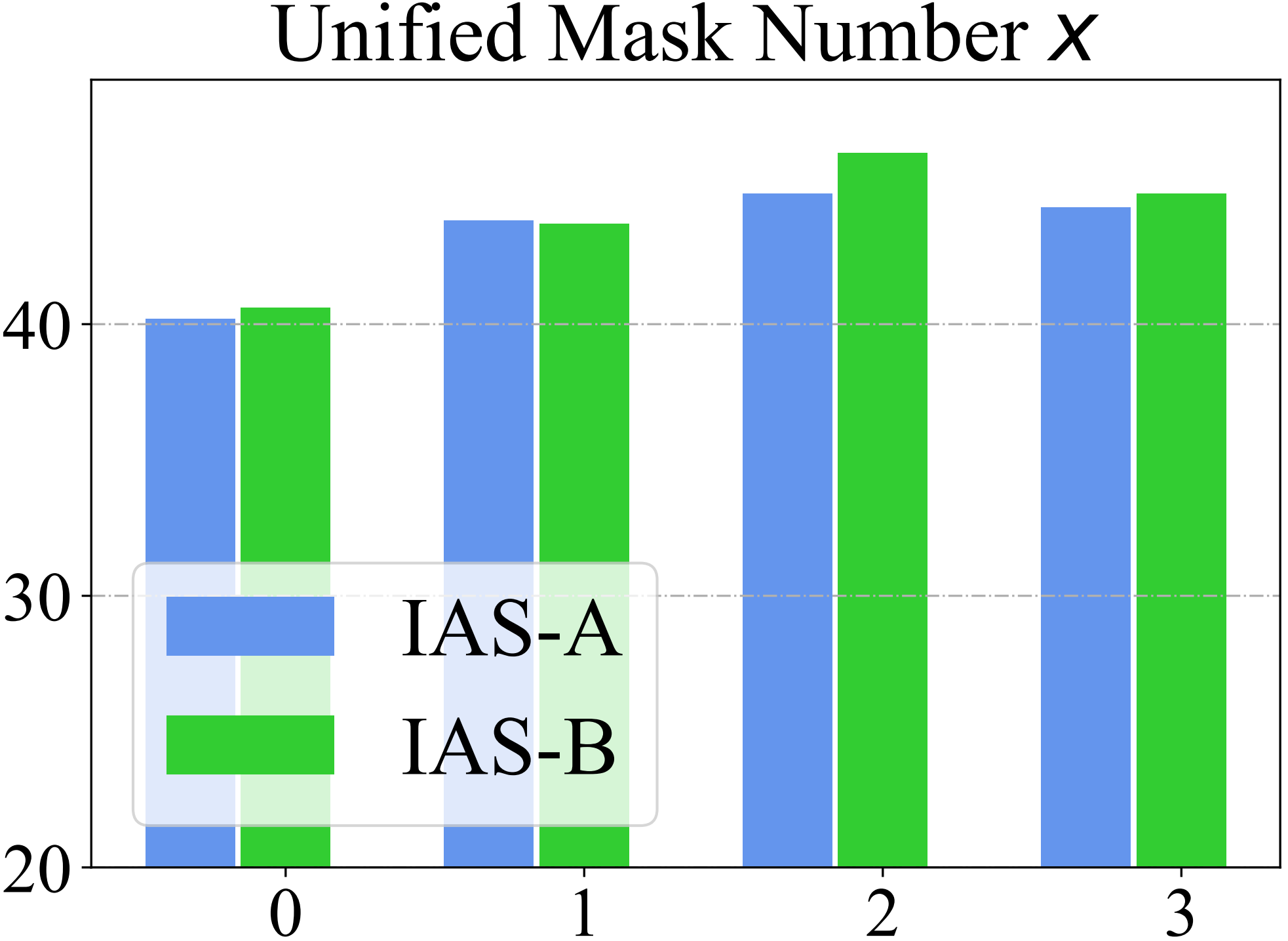}} 
  \scalebox{0.25}{\includegraphics[]{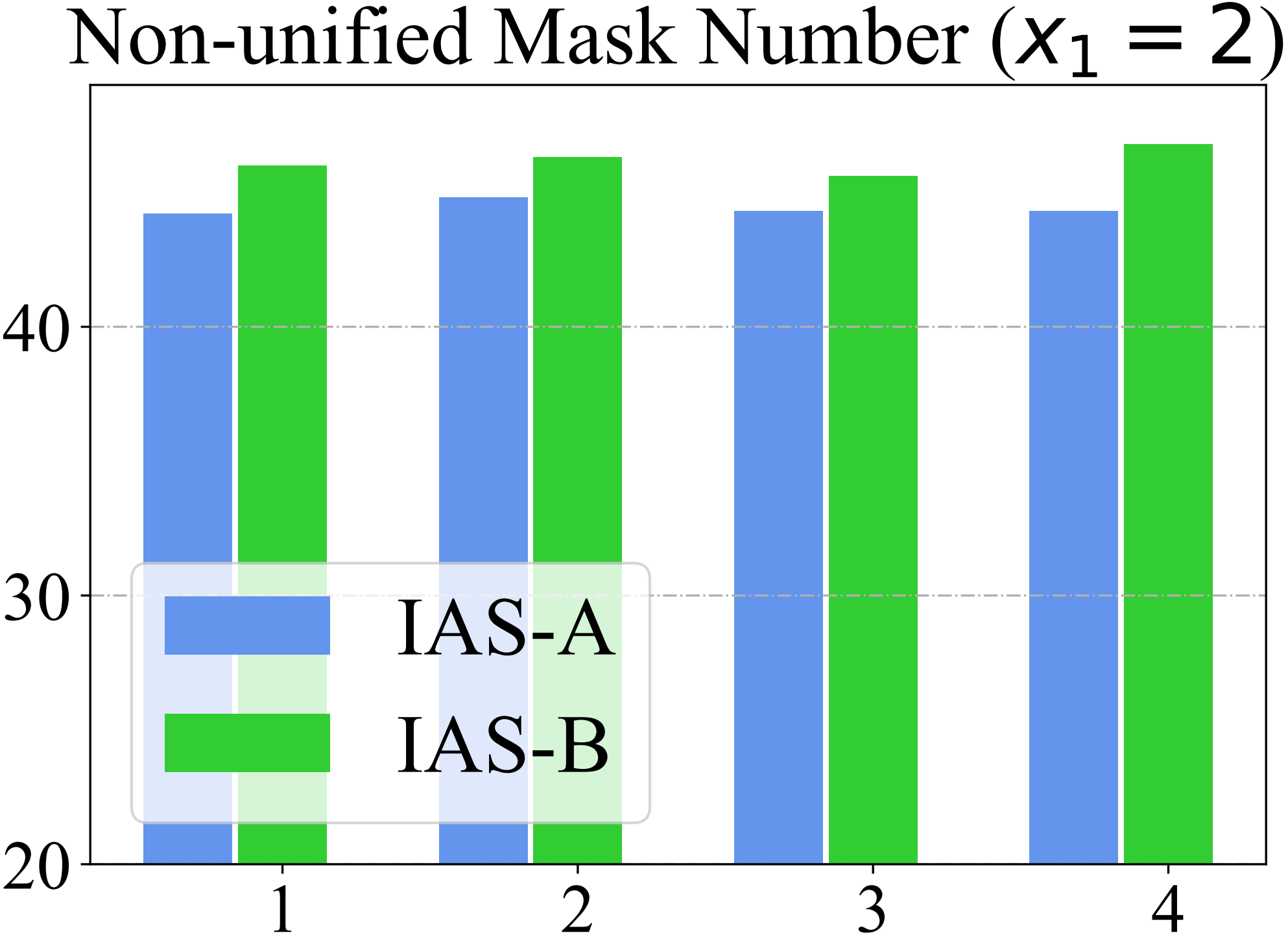}}
  \\
  \scalebox{0.25}{\includegraphics[]{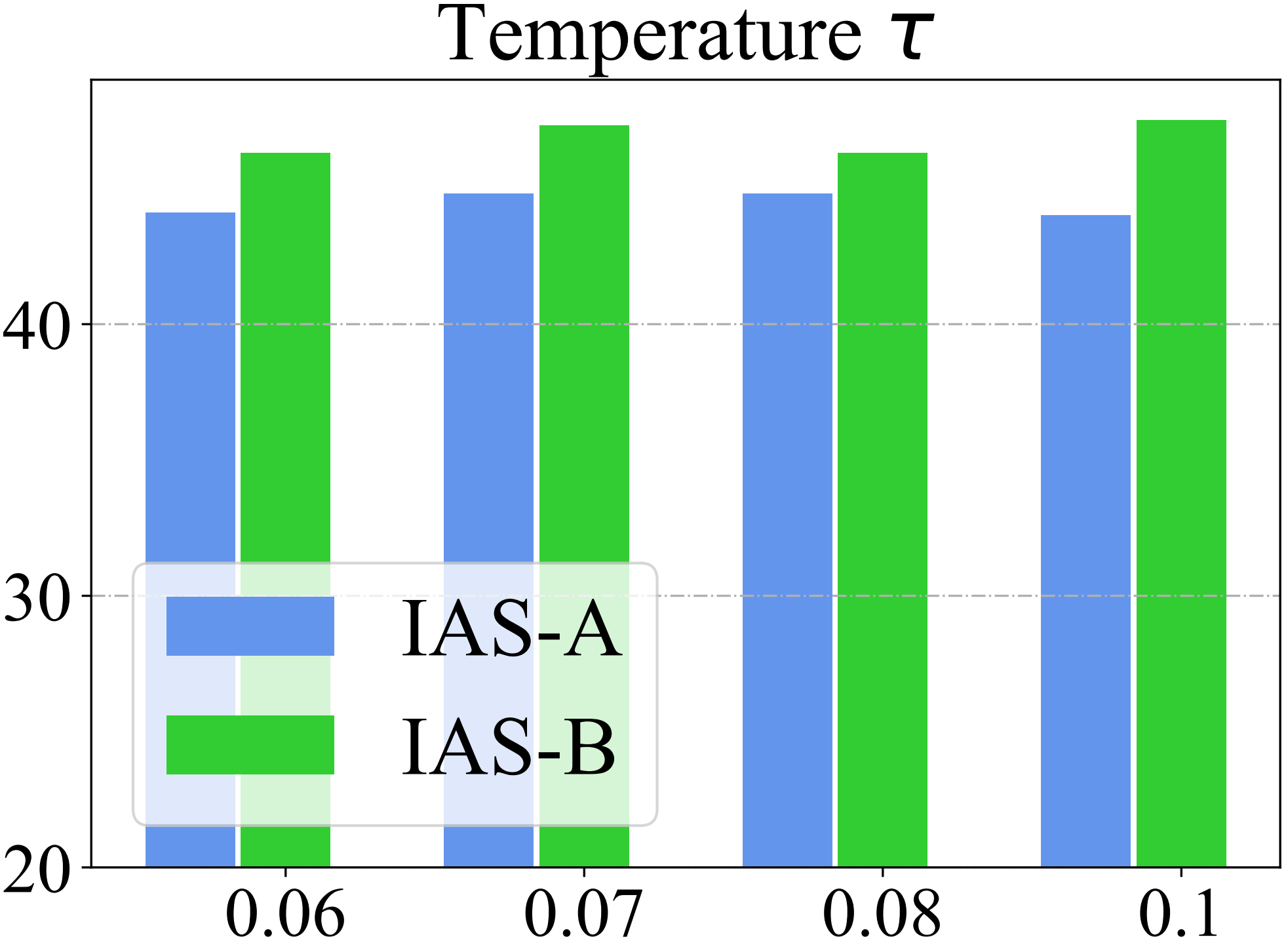}}
      \scalebox{0.25}{\includegraphics[]{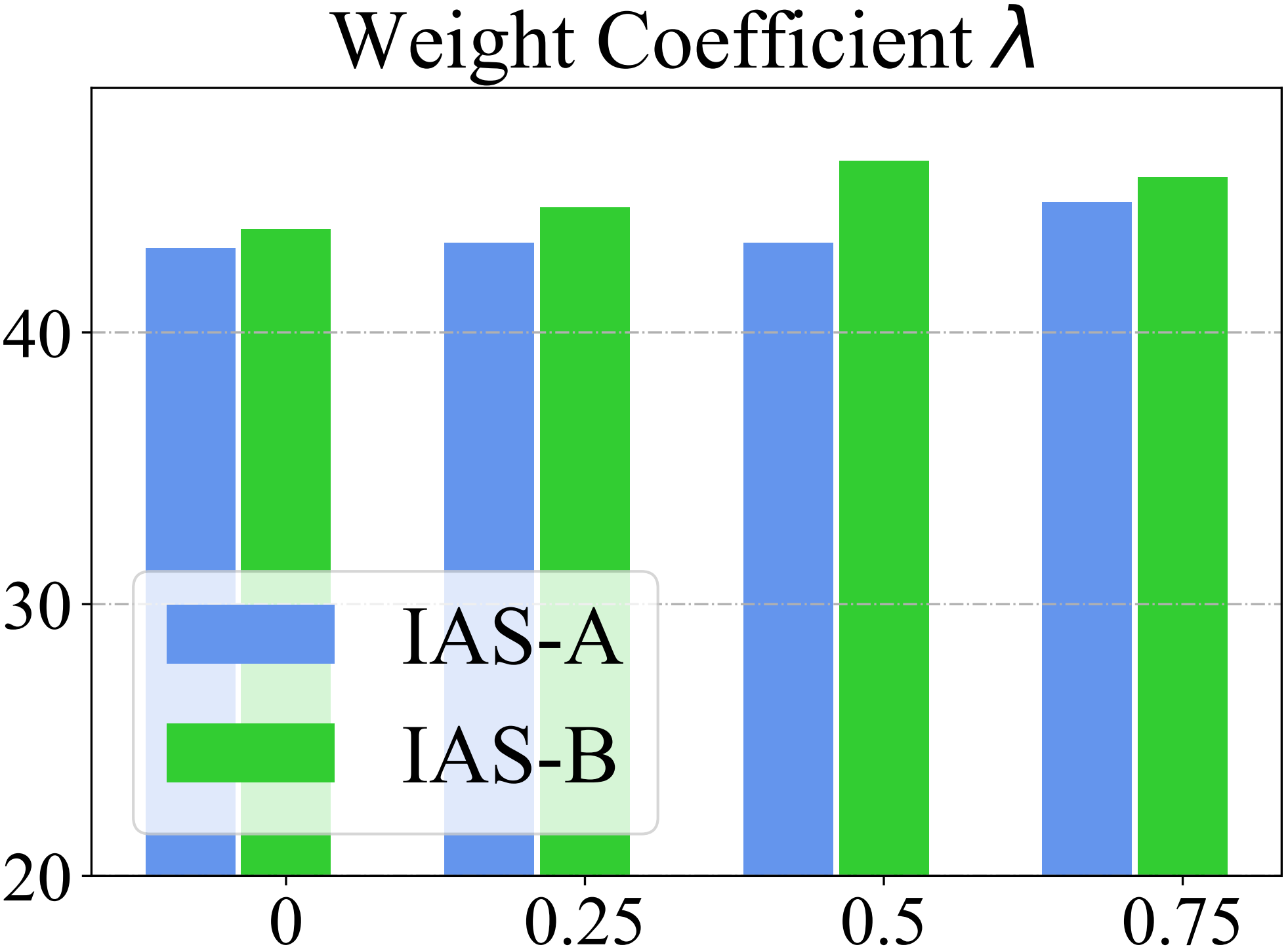}} \\
         \scalebox{0.25}{\includegraphics[]{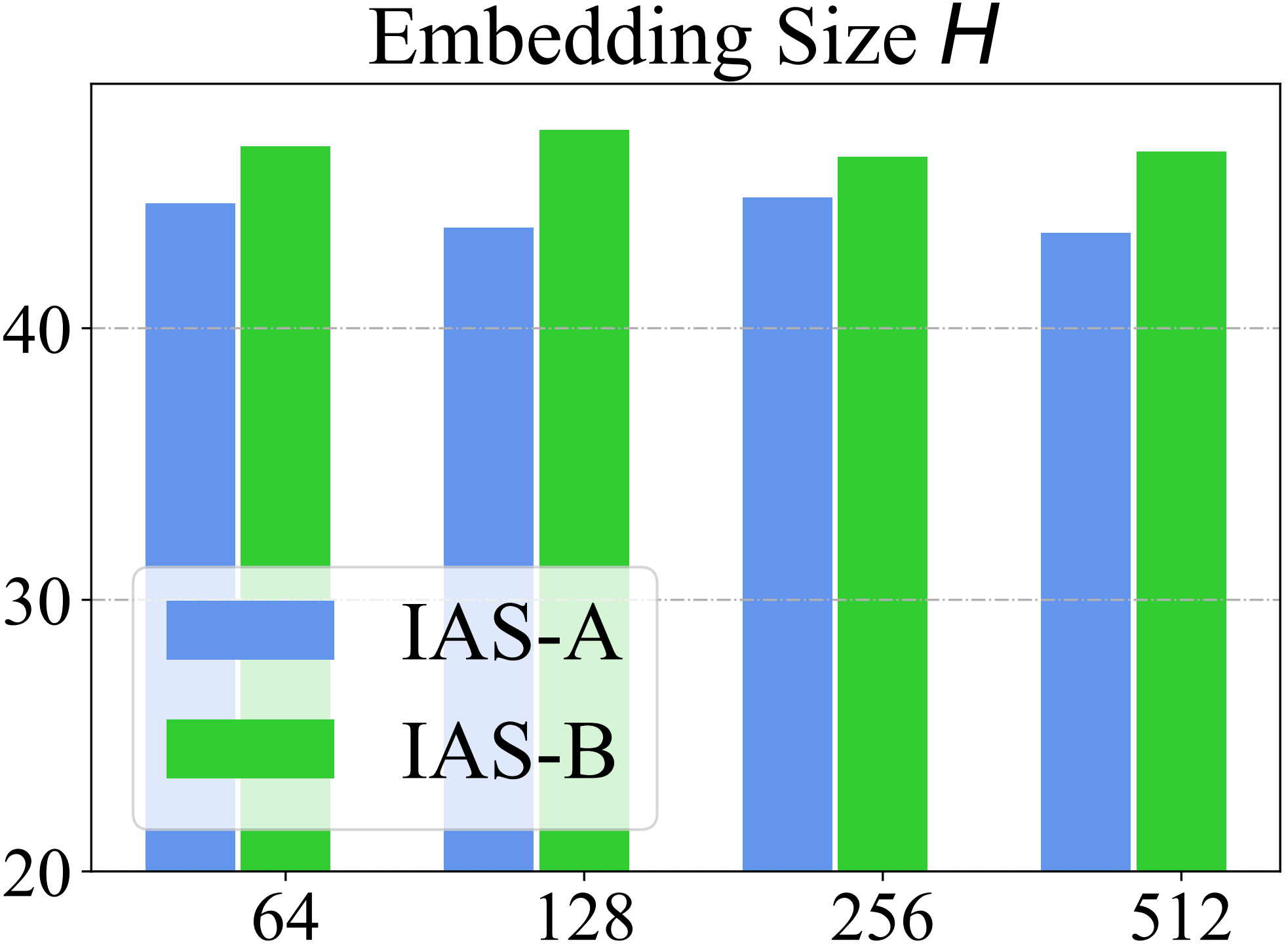}}
         \scalebox{0.25}{\includegraphics[]{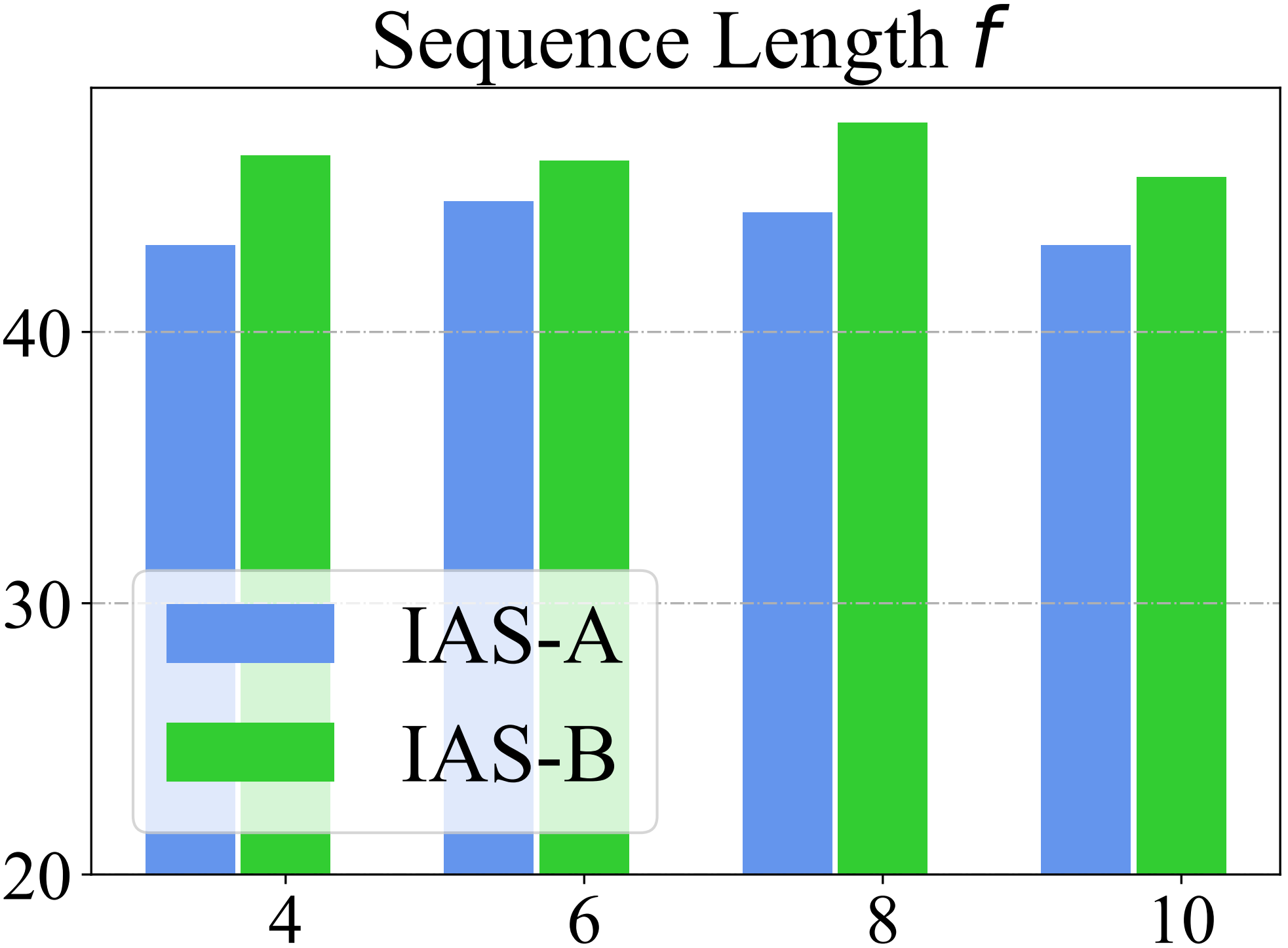}}
    \caption{Top-1 accuracy on IAS-A/B showing effects of hyper-parameters. ``Non-unified Mask Number ($x_{1}=2$)'' denotes using different mask numbers including $x=2$ for subsequence sampling.}
    \label{parameters}
\end{figure}

\section{Further Analysis}
\label{further}
\paragraph{Application to Model-estimated Skeletons.} 
To verify the effectiveness of SimMC when applied to RGB-based scenes with model-estimated 3D skeletons, we utilize pre-trained pose estimation models to extract skeleton data from RGB videos of CASIA-B, and compare the performance of SimMC with representative appearance-based and skeleton-based methods. As shown in Table \ref{CASIA-B}, the proposed SimMC remarkably outperforms state-of-the-art skeleton-based models SM-SGE and SGELA by a distinct margin of $5.6\%$ to $42.5\%$ top-1 accuracy and $0.4\%$ to $8.6\%$ mAP in different conditions, which suggests the stronger ability of our framework on capturing discriminative features from estimated skeleton data for person re-ID. Compared with appearance-based ELF and MLR models that utilize visual features ($e.g.$, colors, textures, and silhouettes) with extra label information, the skeleton-based SimMC also achieves superior performance in all conditions of CASIA-B, which demonstrates its great applicable value and potential for person re-ID under large-scale RGB-based scenarios and more general settings.

\paragraph{Ablation Study.} We conduct ablation study to demonstrate the contribution of each component in our framework. \hc{We adopt 3D coordinates of raw skeleton sequences as the baseline representation for person re-ID.} As reported in Table \ref{ablation}, the model exploiting NPC significantly outperforms the baseline by $9.8$-$47.8\%$ top-1 accuracy and $2.0$-$11.0\%$ mAP. Considering that NPC is a special case of the proposed MPC scheme (see Sec. \ref{MPC_sec}), such results verify the effectiveness of the skeleton prototype contrastive learning in MPC, which can capture highly discriminative features within unlabeled skeleton sequences for the task of person re-ID. Employing the standard MPC scheme with randomly sampled subsequences consistently improves the model performance by up to $3.9\%$ top-1 accuracy and $1.2\%$ mAP on all datasets, which demonstrates that MPC is able to mine more representative key features from skeleton subsequences to perform person re-ID. Finally, incorporating MIC into MPC further improves model performance with $0.8$-$2.5\%$ top-1 accuracy and $0.2\%$-$1.2\%$ mAP gains on different datasets. This justifies our claim that capturing inherent intra-sequence similarity and pattern consistency within sequences could facilitate learning better representations of skeleton sequences for person re-ID.

\paragraph{Discussions.} As shown in Fig. \ref{TNS_comp}, we conduct a t-SNE visualization \cite{van2008visualizing} of representations. The skeleton representations learned by our framework are clustered with higher inter-class separation than AGE and SM-SGE, which suggests that SimMC may learn richer class-related semantics and lower-entropy skeleton representations. We also show effects of different parameters on SimMC in Fig. \ref{parameters}, which indicates that the use of random masks ($x>0$) is the key to the proposed masked contrastive learning, regardless of adopting unified or non-unified mask numbers, while an appropriate fusion ($\lambda>0$) of MIC and MPC facilitates better skeleton representation learning for person re-ID. Our framework with the optimal parameter setting is not sensitive to changes of some parameters such as temperatures $\tau$. More results and proof are provided in the \hc{appendices}.

\section{Conclusion}
In this paper, we propose a simple masked contrastive learning (SimMC) framework to efficiently learn representations of unlabeled skeleton sequences for unsupervised person re-ID. A novel masked prototype contrastive learning (MPC) scheme is devised to cluster the most typical skeleton features of subsequences randomly masked from original sequences, so as to contrast their inherent similarity to learn a discriminative skeleton representation from unlabeled skeletons. To fully exploit inherent relationships between subsequences, we propose a masked intra-sequence contrastive learning (MIC) to learn their similarity and pattern consistency within the sequence for more effective skeleton representations. Our framework outperforms existing state-of-the-art skeleton-based methods and also enjoys high scalability and efficiency to be applied to different models and scenes.

\section*{Ethics Statement}
Person re-ID as an important emerging research topic possesses great value for both academia and industry. However, illegal or improper use of person re-ID technologies could pose serious threat to the public privacy and society security. Therefore, \hc{it should be noted that} all datasets used in our experiments are officially shared by reliable public (IAS-Lab, BIWI, KGBD) or private research agency (KS20, CASIA-B), which have guaranteed that the collecting, processing, releasing, and using of all data are with the consent of participated subjects. For the protection of privacy, all individuals are anonymized with simple identity numbers. Our models and codes must only be used for the purpose of research.

\section*{Acknowledgements}
This work was supported by Alibaba Group through Alibaba Innovative Research (AIR) Program and Alibaba-NTU Singapore Joint Research Institute (JRI), Nanyang Technological University, Singapore.

\bibliographystyle{named}
\bibliography{ijcai22}



\end{document}